\documentclass[10pt,twocolumn,letterpaper]{article}

\usepackage{cvpr}              

\definecolor{cvprblue}{rgb}{0.21,0.49,0.74}
\usepackage[pagebackref,breaklinks,colorlinks,allcolors=cvprblue]{hyperref}

\usepackage{algorithm}
\usepackage{algpseudocode}

\usepackage{booktabs}
\usepackage{multirow}

\usepackage{graphicx}   
\usepackage{array}      
\usepackage{makecell}   
\usepackage{subcaption}

\def\paperID{} 
\def\confName{CVPR}
\def\confYear{2026}

\title{Accelerating Diffusion via Hybrid Data-Pipeline Parallelism \\ Based on Conditional Guidance Scheduling}

\newcommand\blfootnote[1]{%
  \begingroup
  \renewcommand\thefootnote{}\footnote{#1}%
  \addtocounter{footnote}{-1}%
  \endgroup
}

\author{Euisoo Jung \qquad Byunghyun Kim \qquad Hyunjin Kim \qquad Seonghye Cho \qquad Jae-Gil Lee\textsuperscript{*}\\
School of Computing, KAIST  \\
\tt\small \{jyssys, rooknpown, hjkim1228, orangingq, jaegil\}@kaist.ac.kr
}

\begin{document}


\twocolumn[{%
\renewcommand\twocolumn[1][]{#1}%
\vspace{-13mm}
\maketitle

\vspace{-12mm}
\begin{center}
    \captionsetup{type=figure}
    \includegraphics[width=1.00\linewidth]{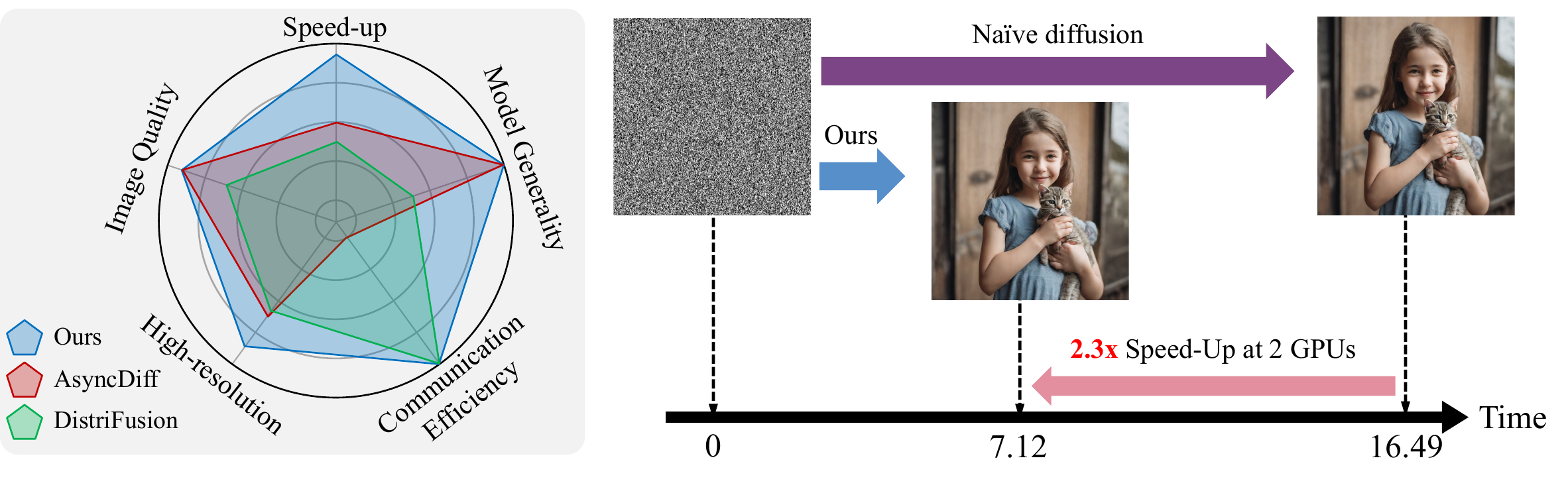}
    \vspace{-12mm}
    \caption{\textbf{Summary of the proposed hybrid data-pipeline parallelism.} Our method consistently outperforms prior distributed approaches across five key aspects: \textit{Speed-up}, \textit{Image Quality}, \textit{Generality}, \textit{High-resolution Synthesis}, and \textit{Communication Cost}, demonstrating robust and balanced acceleration-quality trade-offs.}
    \label{fig:teaser}
\end{center}%
}]

\blfootnote{\hspace{-1.8em}$^*$ indicates corresponding author.}

\begin{abstract}
Diffusion models have achieved remarkable progress in high-fidelity image, video, and audio generation, yet inference remains computationally expensive. Nevertheless, current diffusion acceleration methods based on distributed parallelism suffer from noticeable generation artifacts and fail to achieve substantial acceleration proportional to the number of GPUs. Therefore, we propose a hybrid parallelism framework that combines a novel data parallel strategy, condition-based partitioning, with an optimal pipeline scheduling method, adaptive parallelism switching, to reduce generation latency and achieve high generation quality in conditional diffusion models. The key ideas are to (i) leverage the conditional and unconditional denoising paths as a new data-partitioning perspective and (ii) adaptively enable optimal pipeline parallelism according to the denoising discrepancy between these two paths. Our framework achieves $2.31\times$ and $2.07\times$ latency reductions on SDXL and SD3, respectively, using two NVIDIA RTX~3090 GPUs, while preserving image quality. This result confirms the generality of our approach across U-Net-based diffusion models and DiT-based flow-matching architectures. Our approach also outperforms existing methods in acceleration under high-resolution synthesis settings. Code is available at https://github.com/kaist-dmlab/Hybridiff.
\end{abstract}

\section{Introduction}
\label{sec:intro}

\begin{figure*}[t]
  \centering
  \includegraphics[width=1.00\linewidth]{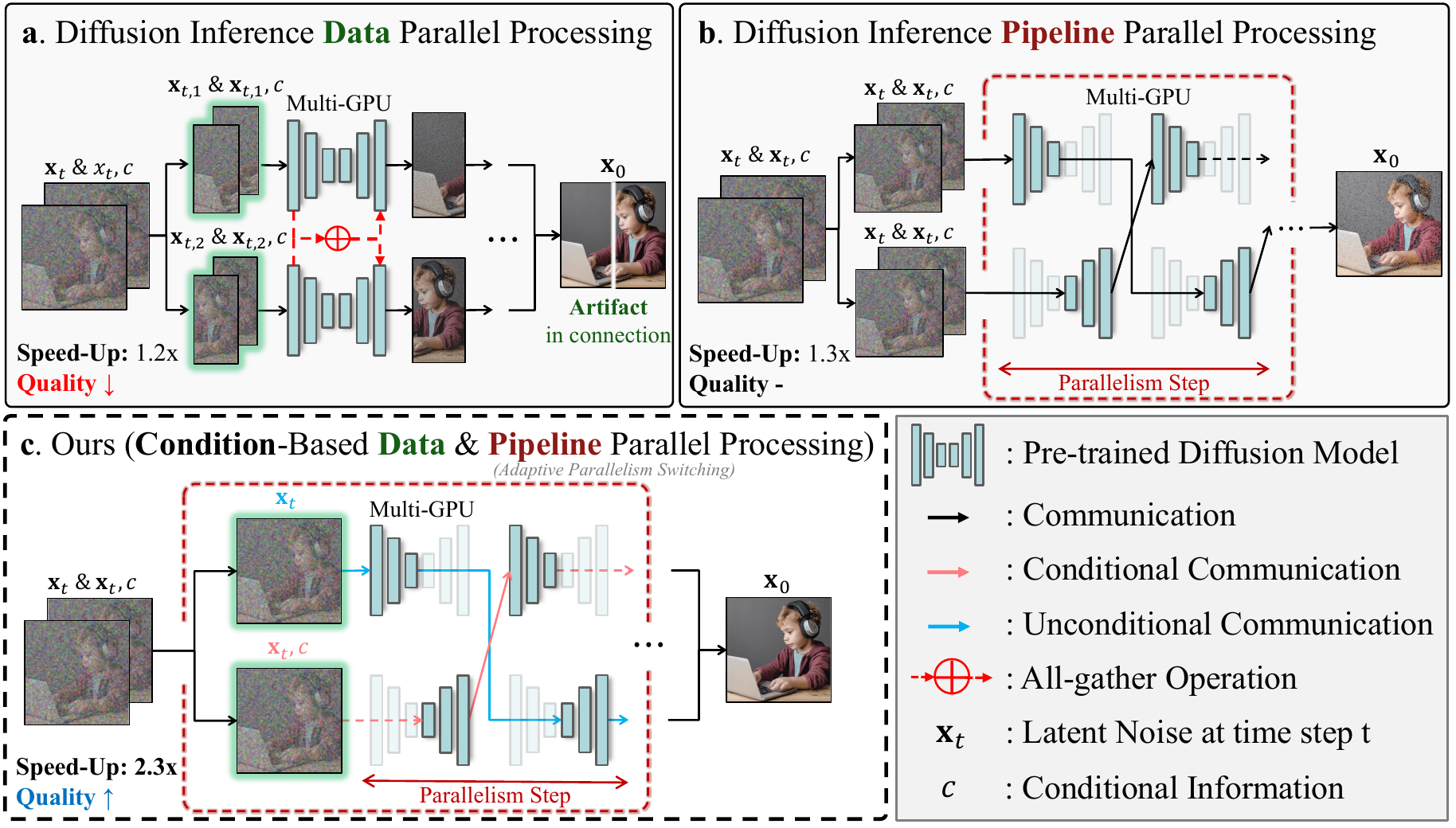}
  \vspace{-7mm}
  \caption{\textbf{Comparison of parallel strategies for diffusion inference.} (a) Patch-based data parallel frameworks suffer from bottlenecks caused by all-gather operations and artifacts at patch boundaries, leading to limited acceleration and quality degradation. (b) Pipeline parallel frameworks incur excessive asynchronous communication overhead and accumulate estimate errors. (c) Our \textbf{hybrid parallelism}, which incorporates {condition}-based data parallelism, adaptively combines both paradigms to achieve high fidelity and fast generation.}
  \label{fig:comparison_overview}
  \vspace{-4.2mm}
\end{figure*}


\vspace{5mm}
Diffusion models have emerged as a powerful family of generative models because of their superior sample quality and broad applicability. However, the inherently iterative nature of diffusion processes, which consists of many denoising steps, leads to significant inference latency and computational bottlenecks. As model sizes continue to scale, these inefficiencies become increasingly limiting, making diffusion inference acceleration a pressing research challenge. Existing approaches have focused mainly on reducing the number of sampling steps\,\cite{kong2021fast, lu2022dpm, lu2025dpm, salimans2022progressive, sauer2024adversarial, yin2024one, xiao2021tackling, yue2023resshift, luo2023latent}, designing optimal architectures\,\cite{li2022efficient, li2023q, zhang2023xformer, li2023snapfusion,zhang2024laptop, yang2023diffusion}, or leveraging mathematical approximations\,\cite{bao2022analytic, liu2022pseudo, zhang2023fast, zhang2023fast, ma2024deepcache, liu2024faster, shih2023parallel}. Yet, these methods often require additional training or fail to deliver strong acceleration in practice, exhibiting a clear trade-off between generation quality and speed.

Distributed parallelism across multiple GPUs offers a promising alternative. Using modern parallel computing resources, one can achieve substantial throughput improvements in diffusion inference without additional training. This direction is especially appealing given the success of distributed strategies in natural language processing, where large-scale language models have already benefited from extensive parallelism research\,\cite{rasley2020deepspeed, shoeybi2019megatron}. As in other domains, distributed parallelism for generative model inference can be broadly classified into \textit{data parallelism} and \textit{pipeline parallelism}\,\cite{li2024distrifusion, chen2024asyncdiff}. Both approaches enhance throughput by distributing either the input data or the model itself across multiple GPUs. 

Representative existing studies include DistriFusion\,\cite{li2024distrifusion} for data parallelism and AsyncDiff\,\cite{chen2024asyncdiff} for pipeline parallelism. In DistriFusion (Figure \ref{fig:comparison_overview}a), an input image is divided into $N$ disjoint patches, and these patches are processed in parallel across $N$ GPUs, where each device independently handles one patch. 
In AsyncDiff (Figure \ref{fig:comparison_overview}b), the entire model is divided into $N$ sequential components, where each component is assigned to a GPU, and the output from the $i$-th GPU is asynchronously fed as the input to the $(i+1)$-th GPU; thus,  AsyncDiff enables pipelined execution across devices.

In theory, each form of parallelism can improve throughput linearly with respect to the number of GPUs, up to an ideal $N\times$speed-up, but in practice, the gains are often sublinear due to communication overhead and synchronization costs. In this paper, we propose a \emph{hybrid} strategy that combines data and model parallelism to further increase the throughput of generative model inference, achieving \emph{beyond-linear scaling} relative to the number of GPUs, while maintaining generation quality. That is, if there are two GPUs, we aim to obtain more than a twofold speed-up without noticeable degradation in output fidelity. In practice, when using two GPUs, data and model parallelism achieved 1.2$\times$ and 1.3$\times$speed-up, respectively, whereas our hybrid approach remarkably achieved a 2.3$\times$speed-up under the same configuration, as shown in Figures \ref{fig:teaser} and \ref{fig:comparison_overview}.

To achieve hybrid parallelism, one could combine the aforementioned representative methods. Specifically, an image is divided into disjoint patches, and each patch is fed into a corresponding model component (not necessarily the first one). As a result, each GPU trains a $1/N$ portion of the model using a $1/N$ portion of an input image. This hybrid approach can potentially achieve beyond-linear scaling; however, it may degrade generation quality for two main reasons. First, since each GPU processes only a portion of the image, artifacts are likely to appear particularly along patch boundaries. Second, this issue is exacerbated by asynchronous communication between model components; that is, errors introduced by asynchronous rather than sequential denoising can worsen the artifacts.

In this paper, we aim to propose and further optimize the hybrid parallelism for diffusion inference from two complementary perspectives: (1) from the data parallelism perspective, transitioning from patch-based partitioning to \emph{condition-based partitioning}; and (2) from the model parallelism perspective, advancing from static parallelism switching to \emph{adaptive parallelism switching}.

\smallskip
\noindent
\textbf{(1) Condition-Based Partitioning.} The main limitation of patch-based partitioning is that each patch represents only a \emph{local} subregion of an image, often leading to boundary artifacts and degraded visual coherence. To address this limitation, we leverage the classifier-free guidance\,(CFG)\,\cite{ho2021classifier}, a technique widely adopted in diffusion models, where the model simultaneously predicts \emph{conditional (prompted)} and \emph{unconditional (unprompted)} noise estimates. This \emph{dual-path} prediction naturally provides a meaningful criterion for data partitioning: as shown in Figure \ref{fig:comparison_overview}c, the conditioned ($\textbf{x}_t,c$) and unconditioned ($\textbf{x}_t$) inputs form two distinct data-parallel paths.
Importantly, unlike patch-based partitioning, each image partition covers the \emph{entire} image, thereby preserving global consistency. Consequently, condition-based partitioning yields improved visual coherence and reduced communication overhead during feature aggregation.

\smallskip
\noindent
\textbf{(2) Adaptive Parallelism Switching.} Because we revise the data partitioning strategy, the pipeline parallelism must also be adapted to align with it. In the early denoising steps, the conditional and unconditional noise estimates differ substantially due to the presence or absence of the condition. Consequently, asynchronous denoising at this stage can lead to divergence between the two paths. To mitigate this issue, we defer the onset of parallel execution until the noise estimates of the two paths become sufficiently similar, beyond the conventional warm-up phase used in prior works (e.g., \cite{chen2024asyncdiff}). Similarly, toward the final denoising steps, the noise estimates from the two paths begin to diverge again; at this point, parallel execution is terminated. The specific switching points between serial and parallel execution are determined automatically based on a novel metric, called the \emph{denoising discrepancy}, which quantifies the difference between the two noise estimates. This \emph{adaptive} switching mechanism effectively improves generation quality by reducing error propagation, while only marginally shortening the duration of parallel processing.


\smallskip
This novel framework demonstrates consistent acceleration not only on conventional denoising diffusion models but also on recent state-of-the-art generative frameworks such as flow matching\,\cite{lipman2023flow}. As long as the model follows a sequential denoising process that allows quantifying the relative influence between conditional and unconditional branches, our framework remains robust and effective. Furthermore, due to the nature of pipeline parallelism, it is not restricted to specific architectures such as {U-Net} or {DiT}, showing strong generality across diverse networks. 

As summarized in Figure~\ref{fig:teaser}, our proposed \emph{hybrid parallelism} achieves superior performance across the five key aspects. In fact, compared to single-GPU inference, our method achieves a \textbf{2.3$\times$speed-up with two GPUs} (i.e., $>2$), while preserving generation fidelity. See Appendix~\ref{app:evaluation_of_hp} and Section~\ref{sec:experiments} for details of Figure~\ref{fig:teaser}.
Finally,
the key contributions are summarized as follows.

\begin{itemize}
    \item \textbf{Hybrid Parallelism Framework for Diffusion Inference.}  
    We introduce a novel diffusion inference parallelism framework that integrates condition-based partitioning and adaptive parallelism switching into a unified hybrid parallelism design.
    
    \item \textbf{Novel Condition-Based Partitioning.}  
    At the data parallelism level, we exploit the intrinsic mechanism of diffusion by decoupling conditional and unconditional branches and performing multi-GPU denoising.
    
    \item \textbf{Adaptive Parallelism Switching.}  
    To align pipeline parallelism with the behavior of conditional guidance, our method adaptively switches to hybrid parallelism framework during inference. Switching points are automatically determined based on the denoising discrepancy between conditional and unconditional estimates, ensuring generation efficiency throughout the denoising process.
    
    \item \textbf{Robustness across Models and Architectures.}  
    Our framework consistently demonstrates strong acceleration and generation quality across various architectures (e.g., U-Net, {DiT}) and recent state-of-the-art generative frameworks, such as flow matching, even under high-resolution synthesis settings.
\end{itemize}

\begin{figure*}[t]
  \centering
  \includegraphics[width=0.95\linewidth]{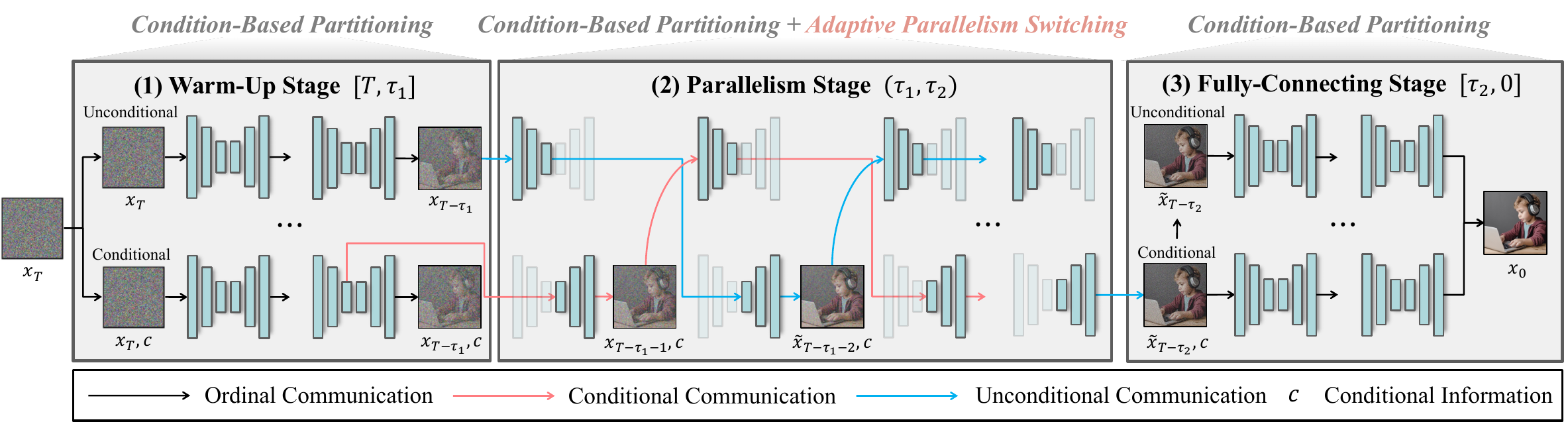}
  \vspace{-2mm}
  \caption{\textbf{Overview of the proposed diffusion inference hybrid parallel framework.} Our method adaptively switches parallelism modes at $\tau_1$ and $\tau_2$, optimizing the trade-off between computational efficiency and consistency of conditional guidance, and demonstrates superior inference acceleration performance while preserving high generation quality.
  }
  \vspace{-5mm}
  \label{fig:method_overview}
\end{figure*}

\section{Related Work}
\label{sec:related_work}

%

\noindent \textbf{\textit{Single}-GPU Diffusion Acceleration.} 
Research on the acceleration of \emph{single}-device diffusion inference can be classified into three categories. The first group focuses on reducing the number of sampling steps required for high-quality generation\,\cite{song2021denoising, kong2021fast, lu2022dpm, lu2025dpm, salimans2022progressive, sauer2024adversarial, yin2024one, xiao2021tackling, zhang2023fast, yue2023resshift, luo2023latent}. These approaches enable fast sampling by either reformulating the reverse process as an ordinary differential equation\,(ODE), distilling multi-step models into fewer steps, or directly predicting the reverse process in latent space. The second group targets model architecture optimization, aiming to reduce computational cost through network compression and efficient design\,\cite{li2022efficient, li2023q, zhang2023xformer, li2023snapfusion, zhang2024laptop, yang2023diffusion}. The third group leverages mathematical and algorithmic strategies, either exploiting the mathematical structure of diffusion processes or reusing intermediate computations to further accelerate inference\,\cite{bao2022analytic, liu2022pseudo, zhang2023fast, ma2024deepcache, wimbauer2024cache, liu2024faster, shih2023parallel}. While these methods reduce single-device inference time, they are inherently limited by the computational capacity of individual GPUs.

\smallskip
\noindent \textbf{\textit{Multi}-GPU Diffusion Acceleration.}
Recent studies have explored various distributed parallelism strategies to accelerate diffusion inference using \emph{multiple} GPUs\,\cite{li2024distrifusion, chen2024asyncdiff, fang2024pipefusion, fang2024xdit, wang2025communication}. DistriFusion\,\cite{li2024distrifusion} introduces a data-parallel approach that divides the input image into independent patches, performing denoising in parallel across GPUs. This work has established a foundational paradigm for parallel diffusion inference. 
Building on this parallelization idea,
AsyncDiff,\cite{chen2024asyncdiff} introduces model parallelism by dividing the U-Net into layer-wise segments and employing a stride-based scheduling strategy to balance parallel execution, achieving a notable reduction in latency.

Subsequently, PipeFusion\,\cite{fang2024pipefusion} and XDiT\,\cite{fang2024xdit} combine patch-level parallelism with transformer-oriented parallelism through ring attention. While additional adaptations such as {CFG}-based data parallelism have been introduced, these methods remain limited to inter-image processing and lack deeper architectural integration. Moreover, transformer-specific schemes such as ring attention exhibit limited scalability and inconsistent performance when applied to general diffusion architectures.
More recently, ParaStep\,\cite{wang2025communication} proposes a reuse-then-predict mechanism that leverages the similarity of noise predictions between adjacent denoising steps. By reusing the noise from previous steps before re-prediction, ParaStep enables inter-step parallelization and significantly reduces communication overhead. However, because early and late diffusion steps exhibit larger discrepancies between adjacent noise states, the reuse mechanism can accumulate errors, leading to potential degradation in image quality or restricted speedup.

\section{Preliminaries}
\label{sec:preliminaries}

\textbf{Denoising Diffusion Model.}  
Let $q(\mathbf{x}_{0})$ denote the data distribution and define a forward noising process by
\[
q(\mathbf{x}_{t}\mid \mathbf{x}_{t-1}) = \mathcal{N}\bigl(\mathbf{x}_{t}; \sqrt{1 - \beta_{t}}\,\mathbf{x}_{t-1}, \beta_{t}\mathbf{I}\bigr),
\]
for $t=1,\dots,T$, with variance schedule $\{\beta_{t}\}$. The model learns a parameterized reverse denoising process,
\[
p_{\theta}(\mathbf{x}_{t-1}\mid \mathbf{x}_{t}) 
= \mathcal{N}\bigl(\mathbf{x}_{t-1}; \mu_{\theta}(\mathbf{x}_{t},t), \Sigma_{\theta}(t)\bigr),
\]
by optimizing the variational lower bound,
\[
\mathcal{L}_{\mathrm{VLB}} = \mathbb{E}_{q}\Bigl[\sum_{t=1}^{T} D_{\mathrm{KL}}\bigl(q(\mathbf{x}_{t-1}\mid\mathbf{x}_{t},\mathbf{x}_{0}) \,\|\, p_{\theta}(\mathbf{x}_{t-1}\mid\mathbf{x}_{t})\bigr)\Bigr].
\]

\noindent \textbf{Classifier-Free Guidance ({CFG}).}  
For conditional generation with a condition $c$, the model is trained to predict the noise $\epsilon_{\theta}(\mathbf{x}_{t},t,c)$ and its unconditional variant $\epsilon_{\theta}(\mathbf{x}_{t},t,\varnothing)$. At inference, the samples follow
\vspace{-1mm}
\[
\epsilon_{\mathrm{cfg}} = \epsilon_{\theta}(\mathbf{x}_{t},c,t) + w\bigl(\epsilon_{\theta}(\mathbf{x}_{t},c,t) - \epsilon_{\theta}(\mathbf{x}_{t},t)\bigr),
\]
where $w>0$ is the guidance scale. The adjusted reverse mean becomes
\vspace{-2mm}
\[
\mu_{\mathrm{cfg}}(\mathbf{x}_{t},t,y) 
= \frac{1}{\sqrt{\alpha_{t}}}\Bigl(\mathbf{x}_{t} - \frac{\beta_{t}}{\sqrt{1-\bar\alpha_{t}}}\,\epsilon_{\mathrm{cfg}}\Bigr).
\]

\noindent \textbf{Flow Matching.}  
Given a target distribution $q(\mathbf{x})$ and base distribution $p_{0}(\mathbf{x})$, flow matching defines an ordinary differential equation,
\vspace{-1mm}
\[
\frac{d\mathbf{x}(t)}{dt} = v(\mathbf{x}(t),t),
\]
where the vector field $v_{\theta}$ is learned by minimizing
\[
\mathcal{L}_{\mathrm{FM}} = \mathbb{E}_{t,\mathbf{x}_{0}\sim q}\Bigl\|v_{\theta}\bigl(\mathbf{x}_{t},t\bigr) - \frac{\mathbf{x}_{t}-\mathbf{x}_{0}}{t}\Bigr\|^{2},
\]
with $\mathbf{x}_{t} = \mathbf{x}_{0} + t\mathbf{e}$ for $\mathbf{e}\sim\mathcal{N}(0,\mathbf{I})$. Sampling proceeds by integrating $\dot{\mathbf{x}}=v_{\theta}(\mathbf{x},t)$ from $t=1$ to $t=0$.

\section{Method}
\label{sec:method}

\subsection{Overview}
\label{sec:framework_overview}
Figure~\ref{fig:method_overview} illustrates the overall process of our proposed hybrid parallelism framework. The input isotropic noise latent $\textbf{x}_T$ is fed simultaneously into two denoising branches: the unconditional path $f_\theta(\textbf{x}_t, t)$ and the conditional path $f_\theta(\textbf{x}_t, c, t)$ guided by a textual prompt $c$. where $f_\theta$ denotes the denoising diffusion network parameterized by $\theta$ (e.g. U-Net, DiT). To exploit both global consistency and conditional fidelity, our framework incorporates two complementary dimensions of parallelism, \textit{condition-based partitioning}, and \textit{adaptive parallelism switching}.

Formally, given the denoising model $f_\theta$, the diffusion inference across $N$ devices can be expressed as
\vspace{-2mm}
\[
\begin{gathered}
\mathbf{x}_{t-1}^{(n)} = f_{\theta^{(n)}}(\mathbf{x}_t^{(n)}, c^{(b_n)}, t), \\[-3pt]
n \in \{1, \dots, N\}, \quad b_n \in \{\text{cond}, \text{uncond}\},
\end{gathered}
\]
where each $\theta^{(n)}$ corresponds to the subset of model parameters assigned to the $n$-th device in the pipeline, reflecting adaptive parallelism switching across different network stages. Meanwhile, $b_n \in \{\text{cond}, \text{uncond}\}$ indicates whether the device $n$ handles the conditional or unconditional branch in condition-based partitioning. Accordingly, each device processes either a conditional input with $c$ or an unconditional input without $c$. This formulation jointly represents both condition-based partitioning and adaptive parallelism switching within a unified diffusion framework.

To further enhance performance, the denoising process is divided into three stages according to the temporal dynamics of conditional influence: (1) \textit{Warm-Up Stage}, where only ordinal communication occurs between conditional and unconditional branches; (2) \textit{Parallelism Stage}, where both branches are executed in parallel with conditional exchange; and (3) \textit{Fully-Connecting Stage}, which merges the two branches for the final refinement. The rationale for this three-phase division and the quantitative criteria for determining the boundary points $\tau_1$ and $\tau_2$ are discussed in Section~\ref{sec:framework_detail} and Section~\ref{sec:method_detail}, respectively.

\begin{figure}[t]
  \centering
  \includegraphics[width=1.0\linewidth]{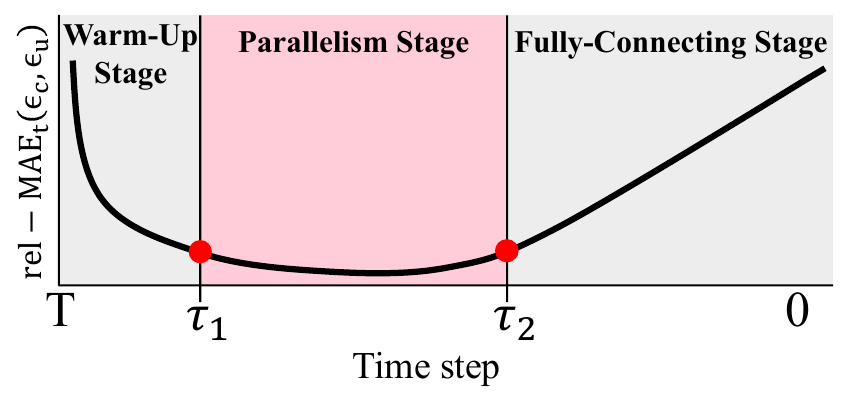}
  \vspace{-8mm}
  \caption{\textbf{Illustration of the $\boldsymbol{\text{rel-MAE}_t(\epsilon_c, \epsilon_u)}$ curve.} The $\text{rel-MAE}_t(\epsilon_c, \epsilon_u)$ value is relatively large before $\tau_1$ and after $\tau_2$, while it converges near zero between them, indicating stable alignment between conditional and unconditional branches during the parallelism phase.}
  \vspace{-5mm}
  \label{fig:rel_mae}
\end{figure}

\subsection{Hybrid Parallel Inference Framework}
\label{sec:framework_detail}

Figure~\ref{fig:rel_mae} illustrates the \underline{rel}ative-\underline{M}ean \underline{A}bsolute \underline{E}rror of the predicted noise (relative-MAE; $\text{rel-MAE}_t(\epsilon_c, \epsilon_u)$) across the three stages of the proposed hybrid parallelism in the denoising diffusion model. To determine when the conditional and unconditional branches should interact or remain independent, we first quantify their \emph{denoising discrepancy} during the denoising process. 

Since conditional and unconditional denoisers contribute differently to generation, with one emphasizing semantic alignment to the text condition and the other stabilizing global structure, it is essential to measure how their noises $\epsilon_c$ and $\epsilon_u$ diverge over time. This discrepancy serves as a key indicator for determining the switching points between serial and parallel execution within our hybrid framework.

The \emph{denoising discrepancy}, namely $\text{rel-MAE}_t(\epsilon_c, \epsilon_u)$, quantifies the difference in noise prediction $\epsilon_t$ between the conditional and unconditional branches at each timestep $t$, (where $\epsilon_c = \epsilon_{\theta}(\textbf{x}_t, c, t)$ and $\epsilon_u = \epsilon_{\theta}(\textbf{x}_t, t)$), and is formulated as
\begin{equation}
\text{rel-MAE}_t(\epsilon_c, \epsilon_u) =
\frac{\mathbb{E}_{\textbf{x},\epsilon}\!\left[\left\lVert \epsilon_{\theta}(\textbf{x}_t, c, t) - \epsilon_{\theta}(\textbf{x}_t, t)\right\rVert_1\right]}
{\mathbb{E}_{\textbf{x},\epsilon}\!\left[\left\lVert \epsilon_{\theta}(\textbf{x}_t, t)\right\rVert_1\right]}.
\label{eq:rel_mae}
\end{equation}
Here, $\epsilon_{\theta}(\textbf{x}_t, c, t)$ and $\epsilon_{\theta}(\textbf{x}_t, t)$ denote the noise components predicted from the conditional and unconditional denoisers, respectively. A larger value indicates a stronger discrepancy between the two branches, reflecting a higher conditional influence on the denoising trajectory at that timestep.

According to the trend of denoising discrepancy shown in Figure~\ref{fig:rel_mae}, which exhibits a U-shaped curve over the entire denoising process, we divide the process into three stages: the \emph{Warm-Up Stage} $[T,\,\tau_1]$, the \emph{Parallelism Stage} $(\tau_1,\,\tau_2)$, and the \emph{Fully Connecting Stage} $[\tau_2,\,0]$. The two parameters, $\tau_1$ and $\tau_2$, define the boundaries between these stages and are automatically determined during the middle of the denoising process. The details of how they are determined are provided in the next section. Intuitively, $\tau_1$ marks the point where the denoising discrepancy ceases to decrease rapidly, while $\tau_2$ indicates the point where it begins to increase.

By measuring denoising discrepancy across 5,000 prompts from the {MS-COCO} 2014 validation set\,\cite{lin2014microsoft}, we observed that the variation of the error between the conditional and unconditional branches exhibits a clear U-shaped trend, as further demonstrated in Appendix~\ref{app:rel_mae_detail}.

We now describe each denoising stage in detail.

\smallskip
\noindent \textbf{(1) Warm-Up Stage [$T,\,\tau_1$].} 
This stage captures the global outline of the generated image. The conditional branch establishes the overall composition from the text prompt, while the unconditional branch stabilizes the coarse structural forms. Since both branches have distinct influences, the denoising discrepancy remains low. Therefore, each branch is processed independently using condition-based partitioning, without adaptive parallelism switching.

\smallskip
\noindent \textbf{(2) Parallelism Stage ($\tau_1,\,\tau_2$).} 
At this phase, the model refines local details within the preformed outline. The conditional and unconditional branches begin to converge, and the denoising discrepancy remains small and stable. To take advantage of this convergence, adaptive parallelism switching is activated, enabling a more powerful acceleration of the denoising process.

\smallskip
\noindent \textbf{(3) Fully-Connecting Stage [$\tau_2,\,0$].} 
In the final phase, fine-grained conditional cues dominate generation. The framework reverts to condition-based partitioning, integrating conditional guidance to reconstruct the final image $\textbf{x}_0$. 

\smallskip
A similar three stages structure has also been observed in previous studies on diffusion conditional guidance\,\cite{kynkaanniemi2024applying}, which further supports the validity of our framework. Building upon this, through this three stages hybrid parallelism framework, our method achieves efficient distributed denoising while preserving generation quality.

The denoising discrepancy, $\text{rel-MAE}_t(\epsilon_c, \epsilon_u)$ can be extended to flow matching models by replacing $\epsilon_{\theta}$ with the predicted velocity $v_{\theta}$. In this case, $\text{rel-MAE}_t(v_c, v_u)$ maintains the same role in quantifying the conditional-unconditional discrepancy over the velocity field.

\subsection{Adaptive Switching via Denoising Discrepancy}
\label{sec:method_detail}

The core of the proposed hybrid parallelism is to dynamically determine the timesteps $\tau_1$ and $\tau_2$ during the real-time denoising process, sequentially switching between the Warm-Up, Parallelism, and Fully-Connecting modes, while constructing a scheduling method based on the previously defined denoising discrepancy.
\smallskip

\noindent \textbf{(1) Determining $\boldsymbol{\tau_1}$.}
For each timestep $t$, we compute denoising discrepancy and calculate the average slope of the most recent $L$ steps by
\vspace{-2mm}
\begin{equation}
G_t = \frac{M_{t} - M_{t-L}}{L}.
\label{eq:slope}
\end{equation}
We then select $\tau'_1 = \min\{t \,|\, 0 \le G_t < g_{\text{slope}}\}$ and constrain it by safety-cap $\tau_{\text{cap}}$. As shown in Appendix~\ref{app:rel_mae_detail}, $\tau_{\text{cap}}$ is defined as the global minimum point of the denoising discrepancy curve, and it serves as an upper bound for $\tau_1$ during automatic selection. The introduction of $\tau_{\text{cap}}$ ensures stability by covering cases where $\tau_1$ is assigned too late or remains undefined due to outlier behaviors, thus maintaining generation quality while maximizing acceleration. 

Consequently, $\tau_1$ is given by $\tau_1 = \min(\tau'_1, \tau_{\text{cap}})$, which marks the end of the warm-up stage where conditional influence stabilizes.

\vspace{1mm}
\noindent \textbf{(2) Determining $\boldsymbol{\tau_2}$.} 
During the parallelism phase, $\epsilon_c$ and $\epsilon_u$ converge to an identical value, making the denoising discrepancy measurement no longer meaningful. Therefore, $\tau_2$ is empirically fixed to a certain number of steps $k$ after $\tau_1$, 
\vspace{-1mm}
\begin{equation}
\tau_2 = \tau_1 + k, \quad k \in \mathbb{N}, \; 1 \le k < T - \tau_1.
\vspace{-2mm}
\label{eq:tau2}
\end{equation}
A larger $k$ extends the parallelism phase, resulting in faster inference but lower generation quality, while a smaller $k$ improves fidelity at the cost of latency. A detailed analysis of quality and speed trade-offs with respect to $k$ is presented in Section~\ref{sec:sensitivity_analysis}, where we empirically verify the optimal balance across various $k$. We also provide the algorithm of Section~\ref{sec:method_detail} in Appendix~\ref{app:algorithm} describes the overall process.

\begin{table*}[t]
    \centering
    \small
    \setlength{\tabcolsep}{4pt}
    \renewcommand{\arraystretch}{1.15}
    
    \resizebox{\linewidth}{!}{
    \begin{tabular}{ccc@{\hspace{-7pt}}c@{\hspace{11pt}}c@{\hspace{11pt}}c@{\hspace{11pt}}c cc cc cc}
    \specialrule{1.6pt}{0pt}{0pt}
    \multirow{2}{*}{\textbf{Base Model}} &
    \multirow{2}{*}{\textbf{Devices}} &
    \multirow{2}{*}{\textbf{Methods}} &
    \multirow{2}{*}{\textbf{Latency (s)} $\downarrow$} &
    \multirow{2}{*}{\textbf{Speed-Up} $\uparrow$} &
    \multirow{2}{*}{\textbf{Comm. (GB)} $\downarrow$} &
    \multicolumn{2}{c}{\textbf{FID} $\downarrow$} &
    \multicolumn{2}{c}{\textbf{LPIPS} $\downarrow$} &
    \multicolumn{2}{c}{\textbf{PSNR} $\uparrow$} \\
    \cmidrule(lr){7-8} \cmidrule(lr){9-10} \cmidrule(lr){11-12}
    & & & & & &
    \makecell{w/ G.\,T.} & \makecell{w/ Orig.} &
    \makecell{w/ G.\,T.} & \makecell{w/ Orig.} &
    \makecell{w/ G.\,T.} & \makecell{w/ Orig.} \\
    \midrule
    
    \multirow{4}{*}{\textbf{\rule{0pt}{3.5ex}Stable Diffusion XL}} &
    \textbf{1} & Original Model &
    16.49 & - & - &
    23.977 & - &
    0.797 & - &
    9.618 & - \\
    \cmidrule(lr){2-12}
    
    & \multirow{3}{*}{\textbf{2}} &
    DistriFusion\,\cite{li2024distrifusion} &
    13.53 & 1.22$\times$ & 0.525 &
    24.164 & 4.864 &
    0.7978 & 0.146 &
    9.597 & 24.634 \\
    \cmidrule(lr){3-12}
    &  &
    AsyncDiff\,\cite{chen2024asyncdiff} (stride=1) &
    12.54 & 1.31$\times$ & 9.830 &
    23.941 & 4.103 &
    0.797 & 0.108 &
    9.586 & 26.387 \\
    \cmidrule(lr){3-12}
    &  &
    \textbf{Ours} ($k$=5) &
    \textbf{7.12} & \textbf{2.31$\times$} & \textbf{0.516} &
    \textbf{23.831} & \textbf{4.100} &
    \textbf{0.796} & \textbf{0.107} &
    \textbf{9.665} & \textbf{26.640} \\
    \specialrule{1pt}{0pt}{0pt}
    
    \multirow{5}[1]{*}{\textbf{\rule{0pt}{4.0ex}Stable Diffusion 3}} &
    \textbf{1} & \rule{0pt}{2.5ex}Original Model &
    19.36 & - & - &
    33.433 & - &
    0.810 & - &
    8.086 & - \\
    \cmidrule(lr){2-12}
    
    & \multirow{4}[1]{*}{\textbf{2}} &
    AsyncDiff\,\cite{chen2024asyncdiff} (stride=1) &
    9.82 & 1.97$\times$ & 1.290 &
    33.379 & 2.032 &
    0.813 & 0.052 &
    8.155 & 27.812 \\
    \cmidrule(lr){3-12}
    &  &
    xDiT-Ring\,\cite{fang2024xdit} &
    14.31 & 1.35$\times$ & 121.646 &
    33.356 & 1.909 &
    0.809 & 0.047 &
    8.085 & 27.857 \\
    \cmidrule(lr){3-12}
    &  &
    Parastep\,\cite{wang2025communication} &
    9.98 & 1.94$\times$ & \textbf{0.032} &
    33.340 & 3.350 &
    0.810 & 0.112 &
    8.091 & 22.917 \\
    \cmidrule(lr){3-12}
    &  &
    \textbf{Ours} ($k$=5) &
    \textbf{9.33} & \textbf{2.07$\times$} & 0.189 &
    \textbf{33.322} & \textbf{1.878} &
    \textbf{0.780} & \textbf{0.046} &
    \textbf{8.229} & \textbf{27.875} \\
    \specialrule{1.6pt}{0pt}{0pt}
    \end{tabular}
    }
    \vspace{-2mm}
    
    \caption{\textbf{Quantitative comparison of parallelism methods on the Stable Diffusion XL and Stable Diffusion 3 models.} We compare our method with existing distributed inference techniques under 1- and 2-GPU. We report both the baseline latency and the corresponding acceleration ratio (\textit{Speed-Up}), Communication efficiency \textit{(Comm.)}, and quantitative metrics assessing generation fidelity. Here, \textit{w/ G.T.} denotes comparison with the ground-truth image, and \textit{w/ Orig.} indicates comparison with the original (single-GPU) model output.}
    \label{tab:main_results}
    \vspace{-4mm}
\end{table*}
\subsection{Theoretical Analysis of Adaptive Switching}
\label{sec:theory}

\noindent \textbf{Analysis of Denoising Discrepancy by Score Decomposition.}
The denoising discrepancy can be theoretically interpreted as a ratio between the conditional information strength and the unconditional data prior. From the score-decomposition perspective\,\cite{song2021score,karras2022elucidating}, can be approximated as
\begin{equation}
\raisebox{3pt}{$
\text{rel-MAE}_t(\epsilon_c, \epsilon_u)
= \frac{\|\epsilon_c - \epsilon_u\|_1}{\|\epsilon_u\|_1}
\approx
\frac{\|\nabla_{\textbf{x}_t}\log p(c|\textbf{x}_t)\|_1}{\|s_u(\textbf{x}_t,t)\|_1}.
$}
\vspace{-2mm}
\label{eq:score_mae}
\end{equation}
$\nabla_{\textbf{x}_t}\log p(c|\textbf{x}_t)$ represents the conditional information strength and $s_u(\textbf{x}_t,t)$ denotes the unconditional score of the data distribution. Consequently, denoising discrepancy measures the relative magnitude between conditional and unconditional components.

In the score formulation of Eq.~(\ref{eq:score_mae}), the unconditional score $s_u(\textbf{x}_t,t)=\nabla_{\textbf{x}_t}\log p(\textbf{x}_t)$ captures the intrinsic structure of the data distribution, while the conditional gradient $\nabla_{\textbf{x}_t}\log p(c|\textbf{x}_t)$ encodes the semantic influence of the conditioning signal $c$. Their relative magnitudes evolve naturally along the diffusion process:

\begin{itemize}
    \item \textbf{Warm-Up Stage:}
    When $\textbf{x}_t$ is close to pure noise, $s_u(\textbf{x}_t,t)$ carries little structural information, whereas $\nabla_{\textbf{x}_t}\log p(c|\textbf{x}_t)$ dominates by guiding the global semantic layout from the prompt, leading to a large denoising discrepancy.

    \item \textbf{Parallelism Stage:}
    As denoising progresses, $s_u(\textbf{x}_t,t)$ reconstructs meaningful local structures and becomes comparable in magnitude to $\nabla_{\textbf{x}_t}\log p(c|\textbf{x}_t)$. This balance satisfies $\|s_u(\textbf{x}_t,t)\|\!\approx\!\|\nabla_{\textbf{x}_t}\log p(c|\textbf{x}_t)\|$, yielding $\frac{d}{dt}\text{rel-MAE}_t(\epsilon_c, \epsilon_u)\!\approx\!0$ and motivates the activation of the parallel inference phase.
    \vspace{1mm}
    \item \textbf{Fully-Connecting Stage:}
    At high SNR, most patterns have been recovered by $s_u(\textbf{x}_t,t)$, while $\nabla_{x_t}\log p(c|\textbf{x}_t)$ contributes to fine-grained alignment and texture refinement, causing a mild increase in denoising discrepancy.
\end{itemize}

This interpretation provides an intuitive explanation of how the relative magnitudes of the conditional and unconditional scores evolve across timesteps, theoretically supporting the three stages proposed (Warm-Up $\rightarrow$ Parallelism $\rightarrow$ Fully Connecting). Detailed derivations of Eq.~(\ref{eq:score_mae}) and the robustness analysis of $\tau_1$ under stochastic denoising noise are shown in Appendix~\ref{app:score_relmae_derivation} and Appendix~\ref{app:robust_determine_tau1}, respectively.

\subsection{Extensibility to Many GPU Configurations}
\label{sec:multi_gpu_configurations}
While the hybrid parallelism framework is optimized for two GPUs, it also scales well to larger even-numbered configurations. We present two extension strategies.

\smallskip
\noindent \textbf{(1) Batch-Level Extension.} In this approach, the model generates $\frac{N}{2}$ samples across $N$ GPUs, where each pair of GPUs produces one image. This structure linearly increases acceleration with the number of GPUs while maintaining near-identical generation quality. However, it is most effective when a large number of samples are generated.

\noindent  \textbf{(2) Layer-Wise Pipeline Extension.} This method extends the adaptive parallelism switching mechanism by dividing the optimal pipeline interval into $N$ layer-wise segments, thereby enabling finer-grained parallel execution across multiple devices. Unlike the batch-level scheme, it can be applied to single-sample generation, though it may incur slightly reduced acceleration efficiency and minor quality degradation due to finer partitioning.

The structures and details of both strategies are provided in Appendix~\ref{app:multi_gpu_overview}. Supporting a degree of parallelism greater than two for a \emph{single} image is deferred to future work.

\section{Experiments}
\label{sec:experiments}

\subsection{Experimental Setup}
\label{sec:experimental_setup}

\textbf{Models.} We evaluate our proposed hybrid parallelism framework on two representative diffusion backbones: Stable Diffusion XL\,({SDXL})\,\cite{podell2024sdxl} and Stable Diffusion 3.0\,({SD3}), a DiT-based flow matching model\,\cite{esser2024scaling}. {SDXL} represents U-Net–based latent diffusion models\,\cite{rombach2022high}, while {SD3} reflects the transformer-based paradigm, demonstrating the generality of our approach.

\noindent \textbf{Datasets.} All experiments are conducted on the MS-COCO Captions 2014 benchmark\,\cite{lin2014microsoft}, using 5,000 validation prompts for text-to-image generation. Generated images are compared against both the ground-truth samples and the single-GPU original model outputs.

\noindent \textbf{Metrics.} We evaluate inference efficiency and generative quality. Latency and speed-up ratio measure acceleration. For quality, we report  FID (Fréchet Inception Distance)\,\cite{heusel2017gans}, LPIPS (Learned Perceptual Image Patch Similarity)\,\cite{zhang2018unreasonable}, and PSNR (Peak Signal-to-Noise Ratio). Lower {FID}/{LPIPS} and higher {PSNR} indicate better generation quality.

For implementation details, please refer to Appendix~\ref{app:implementation_details}.

\begin{figure*}[t]
  \centering
  \includegraphics[width=1.00\linewidth]{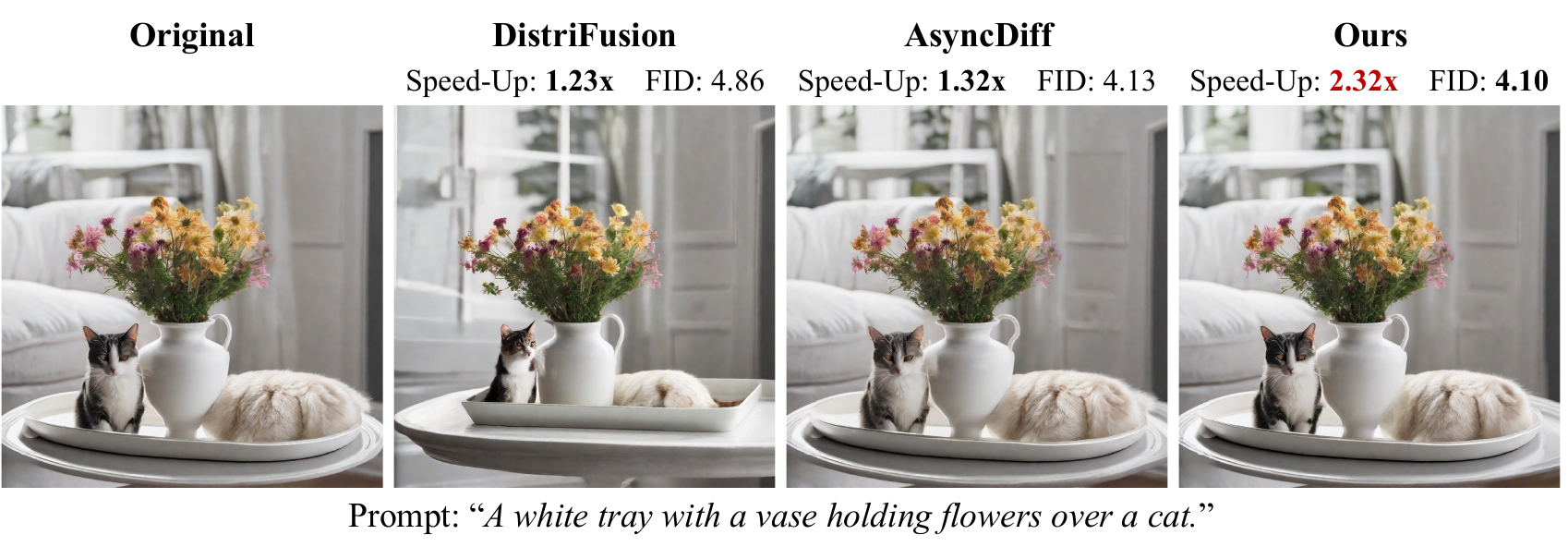}
  \vspace{-8mm}
  \caption{\textbf{Qualitative results of the main experiments.} We compare 1024$\times$1024 image generations from the SDXL model. Our method achieves the best acceleration and FID performance, while producing visuals most similar to the original.}
  \label{fig:main_qualitative}
  \vspace{-5mm}
\end{figure*}

\begin{figure}[t]
  \centering
  \includegraphics[width=1.0\linewidth]{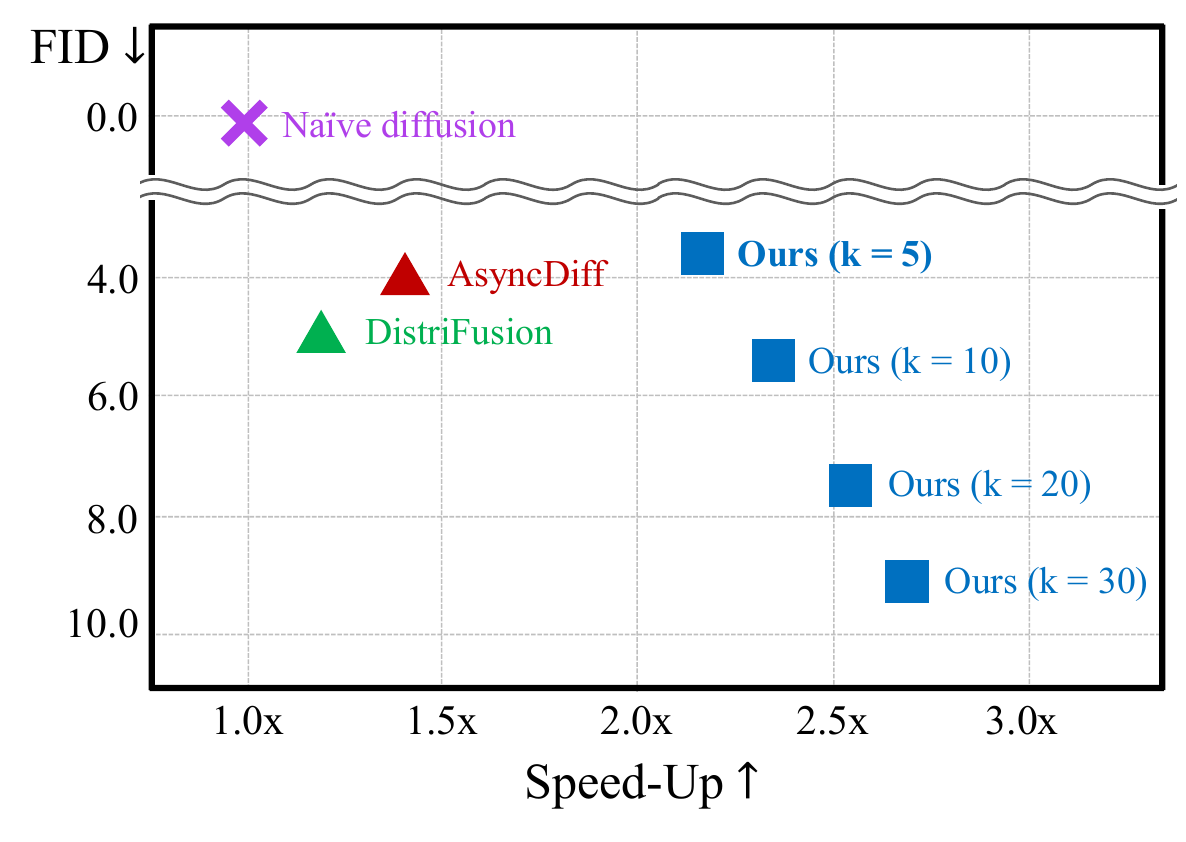}
  \vspace{-8mm}
  \caption{\textbf{Visualization of speed–quality trade-off across different parallelism intervals $\boldsymbol{k}$.} Smaller $k$ values preserve higher fidelity, whereas larger $k$ achieve greater acceleration. Our method consistently dominates prior works across the trade-off frontier. All experiments were conducted on 2 GPUs.}
  \vspace{-5mm}
  \label{fig:k_pareto}
\end{figure}

\subsection{Main Results}
\label{sec:main_results}

\textbf{Quantitative Results.}
Table~\ref{tab:main_results} reports a quantitative comparison across {SDXL} and {SD3} pre-train diffusion models. On {SDXL}, our method achieves a 2.31$\times$ acceleration over the single-GPU baseline while slightly improving image fidelity. Compared to prior distributed inference methods such as DistriFusion,\cite{li2024distrifusion} and AsyncDiff,\cite{chen2024asyncdiff}, our proposed method attains the best speed–quality trade-off with minimal communication overhead. Notably, our communication cost is reduced by 19.6$\times$ compared to AsyncDiff, due to adaptive parallelism switching that dynamically determines optimal parallel intervals to minimize communication cost.

For {SD3}, a DiT-based flow-matching model, our approach not only surpasses earlier distributed frameworks such as DistriFusion and AsyncDiff, but also consistently outperforms more recent baselines, xDiT-Ring and Parastep. It achieves a 2.07$\times$ speed-up with negligible communication cost while maintaining comparable or superior generation quality. These results emphasize our method's strong generality across both U-Net and DiT architectures, achieving generation efficiency.

\vspace{2mm}
\noindent \textbf{Qualitative Results.}
Figure~\ref{fig:main_qualitative} presents qualitative comparisons among distributed inference methods. While DistriFusion and AsyncDiff exhibit boundary artifacts or spatial inconsistency, our method preserves global coherence and fine-grained details similar to the original model. These results confirm that the proposed hybrid parallelism framework maintains high visual fidelity while achieving substantial acceleration. Further results are shown in Appendix~\ref{app:additional_qualitative}.

\begin{table}[t]
  \centering
  \small
  \setlength{\tabcolsep}{2.2pt}
  \renewcommand{\arraystretch}{1.1}
  \resizebox{\columnwidth}{!}{
  \begin{tabular}{@{}cccc@{}}
    \specialrule{1.0pt}{0pt}{0pt}
    \multirow{2}{*}{\textbf{Methods}} &
    \multirow{2}{*}{\textbf{Latency (s)} $\downarrow$} &
    \multirow{2}{*}{\textbf{Speed-Up} $\uparrow$} &
    \textbf{FID} $\downarrow$ \\
    & & & \textbf{(w/ Orig.)} \\
    \midrule
    Original Model          & 16.49 &  -         &  -     \\
    Full Condition-Based Partitioning    &  9.24 &  1.78$\times$ &  \textbf{3.623} \\
    \textbf{Ours (Hybrid Parallelism)} & \textbf{7.12} &  \textbf{2.31$\times$} &  4.100 \\
    \specialrule{1.0pt}{0pt}{0pt}
  \end{tabular}%
  }
  \caption{\textbf{Ablation on hybrid parallel components.} All experiments are conducted on the SDXL model at 1024$\times$1024 resolution, comparing the single-GPU baseline, full condition-based partitioning, and our hybrid parallelism framework.}
  \label{tab:ablation_hybrid_component}
  \vspace{-6mm}
\end{table}

\subsection{Ablation Study}
\label{sec:ablation_study}

\textbf{Ablation on Hybrid Parallel Components.}
Table~\ref{tab:ablation_hybrid_component} analyzes the contribution of each hybrid parallel component. We compare three settings: (1) the original single-GPU model, (2) full condition-based partitioning applied to all denoising steps, and (3) our proposed hybrid parallelism combining both condition-based partitioning and adaptive parallelism switching. Condition-based partitioning achieves a 1.78$\times$ speed-up while maintaining image quality, whereas our hybrid parallelism further improves efficiency to 2.31$\times$ with comparable quality. This demonstrates that the addition of the pipeline component maximizes generation acceleration while minimizing quality degradation. Consequently, the proposed framework effectively integrates the advantages of condition-based partitioning and adaptive parallelism switching.

\vspace{-1mm}
\subsection{Sensitivity Analysis}
\label{sec:sensitivity_analysis}
\noindent \textbf{Impact of Different $k$ Values.}
As shown in Figure~\ref{fig:k_pareto}, the parallelism interval $k$ clearly reveals a speed–quality trade-off: smaller $k$ values preserve higher fidelity, while larger $k$ values yield faster generation. An appropriate balance is observed at $k{=}5$, achieving both strong quality and acceleration. Moreover, the interval $k$ can be flexibly chosen by practitioners to adjust the trade-off between efficiency and fidelity. Quantitative results are summarized in Appendix~\ref{app:effect_of_k_sec}, and qualitative comparisons across different $k$ values are provided in Appendix~\ref{app:qualitative_different_k}.

\begin{figure}[t]
  \centering
  \includegraphics[width=1.0\linewidth]{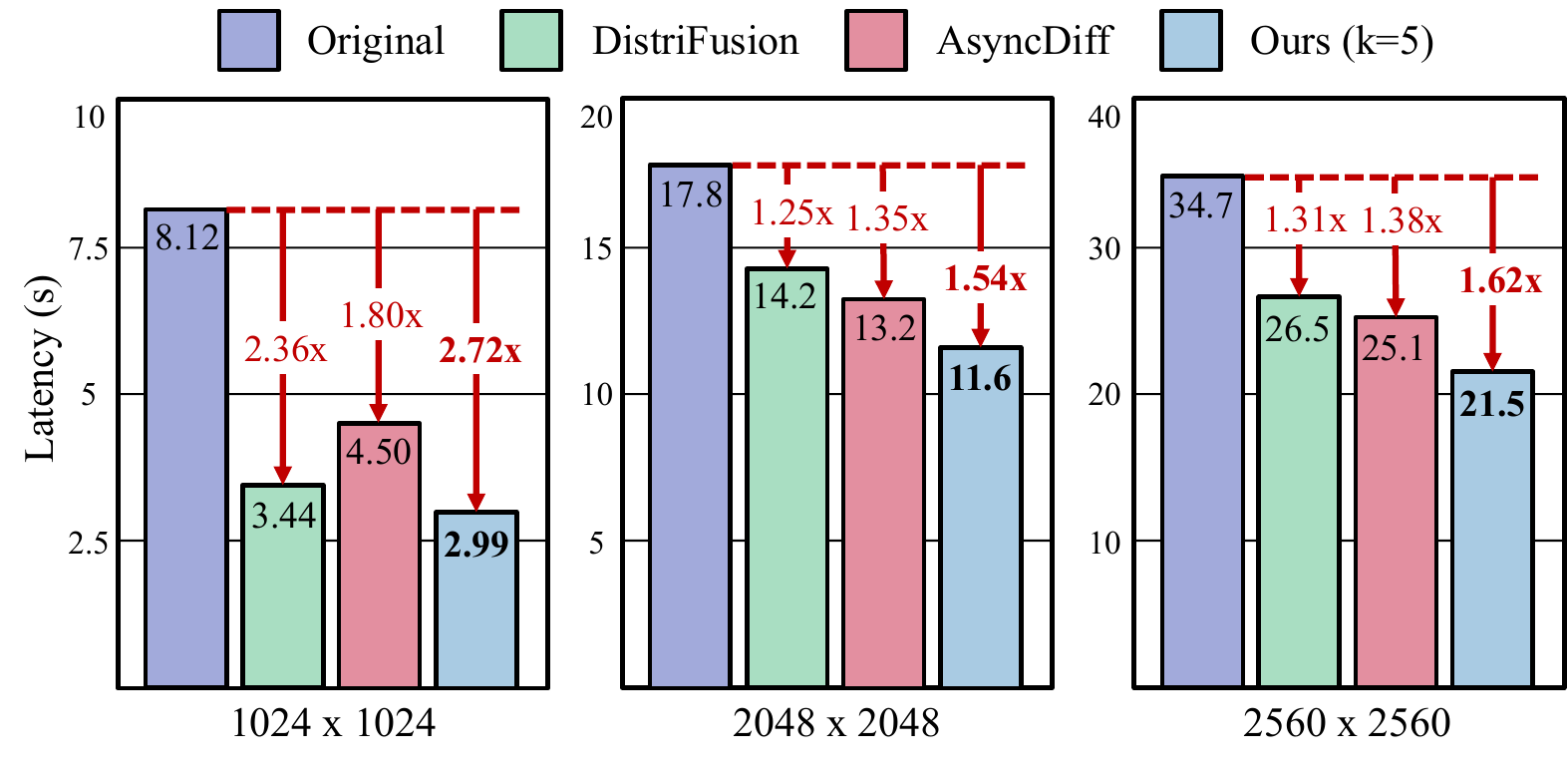}
  \vspace{-9mm}
  \caption{\textbf{Comparison of high-resolutions tasks.} We compare different parallel inference methods on the SDXL model using NVIDIA H200 GPUs across 1024$\times$1024, 2048$\times$2048, and 2560$\times$2560 high-resolutions.}
  \vspace{-6mm}
  \label{fig:high_resolution}
\end{figure}

\smallskip
\noindent \textbf{High-Resolution Generation.}
As shown in Figure~\ref{fig:high_resolution}, our method consistently achieves superior acceleration over existing distributed inference frameworks across increasing resolutions. On the SDXL model using NVIDIA H200 GPUs, our hybrid parallelism attains up to 2.72$\times$ speed-up at 1024$\times$1024, 1.54$\times$ speed-up at 2048$\times$2048, and 1.62$\times$ speed-up at 2560$\times$2560, demonstrating strong scalability for high-resolution image generation.

\vspace{-2mm}

\section{Conclusion}
\label{sec:conclusion}
In this paper, we introduced a hybrid parallelism framework for diffusion inference that integrates condition-based partitioning with adaptive parallelism switching. Guided by the denoising discrepancy criterion, the method adaptively switches between parallelism modes to minimize redundant communication. It achieves $2.31\times$ and $2.07\times$ latency reductions on {SDXL} and {SD3}, respectively, while preserving fidelity. We also generalize across {U-Net} and {DiT} architectures, providing a unified parallelism paradigm for scalable multi-GPU diffusion inference.


{
    \small
    \bibliographystyle{ieeenat_fullname}
    \bibliography{main}
}

\appendix
\clearpage
\setcounter{page}{1}
\maketitlesupplementary


\section{Evaluation of Hybrid Parallelism}
\label{app:evaluation_of_hp}

\begin{table}[t]
  \centering
  \footnotesize
  \setlength{\tabcolsep}{2.5pt}
  \renewcommand{\arraystretch}{1.05}
  \resizebox{\columnwidth}{!}{%
  \begin{tabular}{cccccc}
    \specialrule{1.0pt}{0pt}{0pt}
    \textbf{Methods} & \textbf{Speed-Up} $\uparrow$ & \textbf{\makecell{Image\\Quality}} $\uparrow$ &
    \textbf{\makecell{Model\\General.}} $\uparrow$ & \textbf{\makecell{High-res\\Synth.}} $\uparrow$ &
    \textbf{\makecell{Comm.\\Efficiency}} $\uparrow$ \\
    \midrule
    Distrifusion  & 2.5 & 3.5 & 2.5 & 3.3 & 5.0 \\
    AsyncDiff     & 3.0 & 4.5 & 5.0 & 3.5 & 1.0 \\
    \textbf{Ours} & \textbf{4.7} & \textbf{4.5} & \textbf{5.0} & \textbf{4.4}  & \textbf{5.0} \\
    \specialrule{1.0pt}{0pt}{0pt}
  \end{tabular}
  }
  \caption{\textbf{Quantitative metrics comparison across five evaluation aspects.}
  Scores are normalized to a 5-point scale. Higher values (\,$\uparrow$\,) indicate better performance.}
  \vspace{-3mm}
  \label{tab:eval_hybrid_parallelism}
\end{table}

\noindent\textbf{Evaluation Protocol.} 
All scores are computed based on a 5-point scale unified min--max scaling scheme, where the normalized values are re-centered around an average score of 3. Specifically, each metric is assessed as follows:

\smallskip
\begin{itemize}
    \item \textbf{Speed-Up.} 
    We measure the relative acceleration ratio with respect to the SDXL baseline latency in Table~\ref{tab:main_results}. The measured latencies are 13.53secs for DistriFusion, 12.54secs for AsyncDiff, and 7.12secs for our method.

    \item \textbf{Image Quality.} 
    We evaluate image quality using FID scores reported in Table~\ref{tab:main_results} from the main results of SDXL. The reported FID values are 4.864 for DistriFusion, 4.103 for AsyncDiff, and 4.100 for our method.

    \item \textbf{Model Generality.} 
    We assign scores based on architecture compatibility. Each model receives 2.5 points for supporting U-Net and an additional 2.5 points for DiT support, resulting in scores of 2.5 for DistriFusion, 5 for AsyncDiff, and 5 for Ours.

    \item \textbf{High-resolution Synthesis.} 
    The score reflects both high-resolution generation capability and inference latency. According to the results in Section~\ref{sec:sensitivity_analysis} High-Resolution Generation, all three methods successfully generate three target resolutions. The corresponding average latencies are 14.73secs for DistriFusion, 14.27secs for AsyncDiff, and 11.99secs for Ours.

    \item \textbf{Communication Efficiency.} 
    We evaluate the communication efficiency based on the measured inter-GPU data transfer communication volume in the SDXL multi-GPU setting reported in Table~\ref{tab:main_results} from the main results. The measured communication volumes are 0.525~GB for DistriFusion, 9.830~GB for AsyncDiff, and 0.516~GB for our method.

\end{itemize}

\newpage

\section{Empirical Visualization of Denoising \\ Discrepancy}
\label{app:rel_mae_detail}

\begin{figure}[t]
  \centering
  \vspace{-4mm}
  \includegraphics[width=0.88\linewidth]{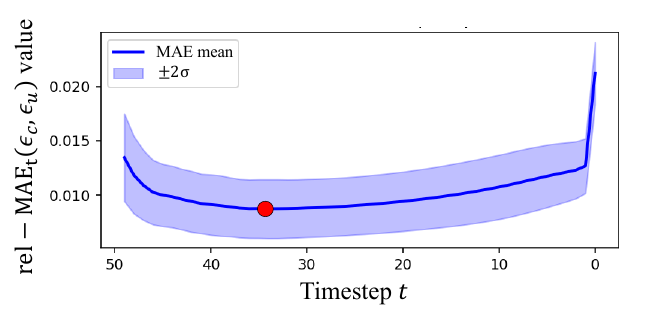}
  \vspace{-4mm}
  \caption{Empirical visualization of denoising discrepancy curve.}
  \vspace{-2mm}
  \label{fig:rel_mae_curve}
\end{figure}

\noindent Figure~\ref{fig:rel_mae_curve} illustrates the average denoising discrepancy ($\text{rel-MAE}_t(\epsilon_c, \epsilon_u)$) value measured during the denoising process based on 5,000 prompts from the {MS-COCO} 2014\,\cite{lin2014microsoft} validation set. The shaded region represents the $\pm2\sigma$ range, and the denoising model used is Stable Diffusion XL. 
The red dot denotes $\tau_{\text{cap}} = \underset{t}{\mathrm{argmin}}\,\text{rel-MAE}_t(\epsilon_c, \epsilon_u)$, which is employed as a safety-cap in the main method.

\section{Adaptive Parallelism Switching Algorithm}
\label{app:algorithm}

\vspace{-3.2mm}
\begin{algorithm}[h]
\caption{Adaptive Parallelism Switching \\ \hspace*{5.3em} via Denoising Discrepancy}
\label{alg:main_alg}
\begin{algorithmic}[1]
\Require latent noise $\textbf{x}_t$, prompt $c$, steps $T$, window $L$, \hspace*{2.3em} slope threshold $g$, safety-cap $\tau_{\text{cap}}$, interval $k$
\State $\tau_1, \tau_2 \gets \varnothing$
\For{$t = T, T\!-\!1, \ldots, 1$}
    \State $\epsilon_c, \epsilon_u \gets \epsilon_\theta(\textbf{x}_t, c, t),~\epsilon_\theta(\textbf{x}_t, t)$
    \State $M_t \gets \dfrac{\mathbb{E}_{x, \epsilon}\|\epsilon_c - \epsilon_u\|_1}{\mathbb{E}_{x, \epsilon}\|\epsilon_u\|_1}$ \Comment{$\text{rel-MAE}_t(\epsilon_c, \epsilon_u)$}
    \State $G_t = \frac{M_t - M_{t-L}}{L}$
    \State \textbf{if} $\tau_1 = \varnothing$ \textbf{ and } $t > L$ \textbf{ and } $0 \le \textstyle G_t < g$ \textbf{then}
        \State \, \, \quad $\tau_1 \gets \min(t,\, \tau_{\text{cap}})$;\quad $\tau_2 \gets \tau_1 + k$
    \State \textbf{Denoise:}
    \If{$t \geq \tau_1$}
        \State \textsc{Warm-Up}
    \ElsIf{$t > \tau_2$}
        \State \textsc{Parallelism}
    \Else
        \State \textsc{Fully-Connecting}
    \EndIf
    \State $x_{t-1} \gets \textsc{Step\;Denoise}(\textbf{x}_t, \epsilon_c, \epsilon_u, t)$
\EndFor
\State \Return $x_0,~(\tau_1, \tau_2)$
\end{algorithmic}
\end{algorithm}

\section{Derivation of Score-Based Interpretation of Denoising Discrepancy}
\label{app:score_relmae_derivation}

The denoising discrepancy($\text{rel-MAE}_t(\epsilon_c, \epsilon_u)$) criterion in Eq.~\eqref{eq:score_mae} can be theoretically derived from the score decomposition of diffusion models. Following the $\epsilon$-parameterization of score-based generative modeling \,\cite{song2021score,karras2022elucidating}, the preconditioned score can be expressed as
\begin{equation}
s_\theta(\textbf{x}_t,t) \approx -\,\frac{\epsilon_\theta(\textbf{x}_t,t)}{\sigma_t},
\label{eq:score_eps}
\end{equation}
where $\sigma_t$ denotes the noise standard deviation at timestep $t$.
According to Bayes' rule, the conditional score function can be decomposed as
\begin{equation}
s_c(\textbf{x}_t,t) = s_u(\textbf{x}_t,t) + \nabla_{\textbf{x}_t}\log p(c|\textbf{x}_t),
\label{eq:score_decomp}
\end{equation}
where $s_u(\textbf{x}_t,t)$ is the unconditional data score, and $\nabla_{\textbf{x}_t}\log p(c|\textbf{x}_t)$ denotes the conditional information flow \,\cite{ho2021classifier}. Substituting Eq.~(\eqref{eq:score_eps}) into Eq.~(\eqref{eq:score_decomp}) yields
\begin{equation}
\epsilon_c(\textbf{x}_t,t) - \epsilon_u(\textbf{x}_t,t)
\propto \sigma_t\,\nabla_{\textbf{x}_t}\log p(c|\textbf{x}_t),
\label{eq:eps_relation}
\end{equation}
which implies that the difference between conditional and unconditional
denoiser outputs corresponds to the conditional gradient scaled by $\sigma_t$. Therefore, the rel-MAE at each timestep $t$ can be approximated as
\begin{equation}
\mathrm{rel\text{-}MAE}_t
=\frac{\|\epsilon_c-\epsilon_u\|_1}{\|\epsilon_u\|_1}
\approx
\frac{\|\nabla_{\textbf{x}_t}\log p(c|\textbf{x}_t)\|_1}{\|s_u(\textbf{x}_t,t)\|_1}.
\label{eq:score_relmae_appendix}
\end{equation}
This formulation reveals that $\text{rel-MAE}_t(\epsilon_c, \epsilon_u)$
quantifies the relative magnitude between the conditional information and the unconditional data prior—forming the theoretical basis for the main method equation (Eq.~\eqref{eq:score_mae}).

\section{Robustness of Determine $\boldsymbol{\tau_1}$ under \\ Stochastic Denoising Noise}
\label{app:robust_determine_tau1}

Diffusion inference is a stochastic denoising process; predicted noises $\epsilon_\theta(\textbf{x}_t)$ are subject to random sampling. Consequently, the observed $\{M_t\}$ fluctuates slightly, and $G_t\!\approx\!0$
may appear prematurely. To ensure robust detection, we define a finite-difference slope by
\begin{equation}
    G_t=\frac{M_t-M_{t-L}}{L},
\end{equation}
which smooths out stochastic perturbations across $L$ timesteps.
The stability of $G_t$ can be theoretically justified by Hoeffding’s inequality:
\begin{equation}
\Pr(|G_t-\mathbb{E}[G_t]|>\delta)
\le 2\exp\!\Big(-\frac{2L\delta^2}{(b-a)^2}\Big).
\end{equation}
Here, $L$ denotes the window length used to compute the moving-average slope, $\delta$ represents the allowable deviation from the expected slope $\mathbb{E}[G_t]$, and $a,b$ correspond to the minimum and maximum possible range of the observed $\text{rel-MAE}_t(\epsilon_c, \epsilon_u)$ values, typically normalized within $[0,1]$. 

As $L$ increases, the variance of the estimated slope decreases, and the probability of false detection decreases exponentially. showing that larger $L$ exponentially reduces false-alarm probability.

Empirically, $L$ and $g_{\text{slope}}$, which are also established in our experiments, lie within a stable regime due to strong autocorrelation of $\text{rel-MAE}_t(\epsilon_c, \epsilon_u)$ sequences. Thus, $\tau_1$ can be reliably detected as
the earliest timestep satisfying $0\!\le\!G_t\!<\!g_{\text{slope}}$
and $t\!\le\!\tau_{\text{cap}}$.

 
\section{Extensibility to Many GPU Configurations Structures}
\label{app:multi_gpu_overview}
\begin{figure*}[t]
  \centering

  \begin{subfigure}[h]{0.9\linewidth}
    \centering
    \includegraphics[width=\linewidth]{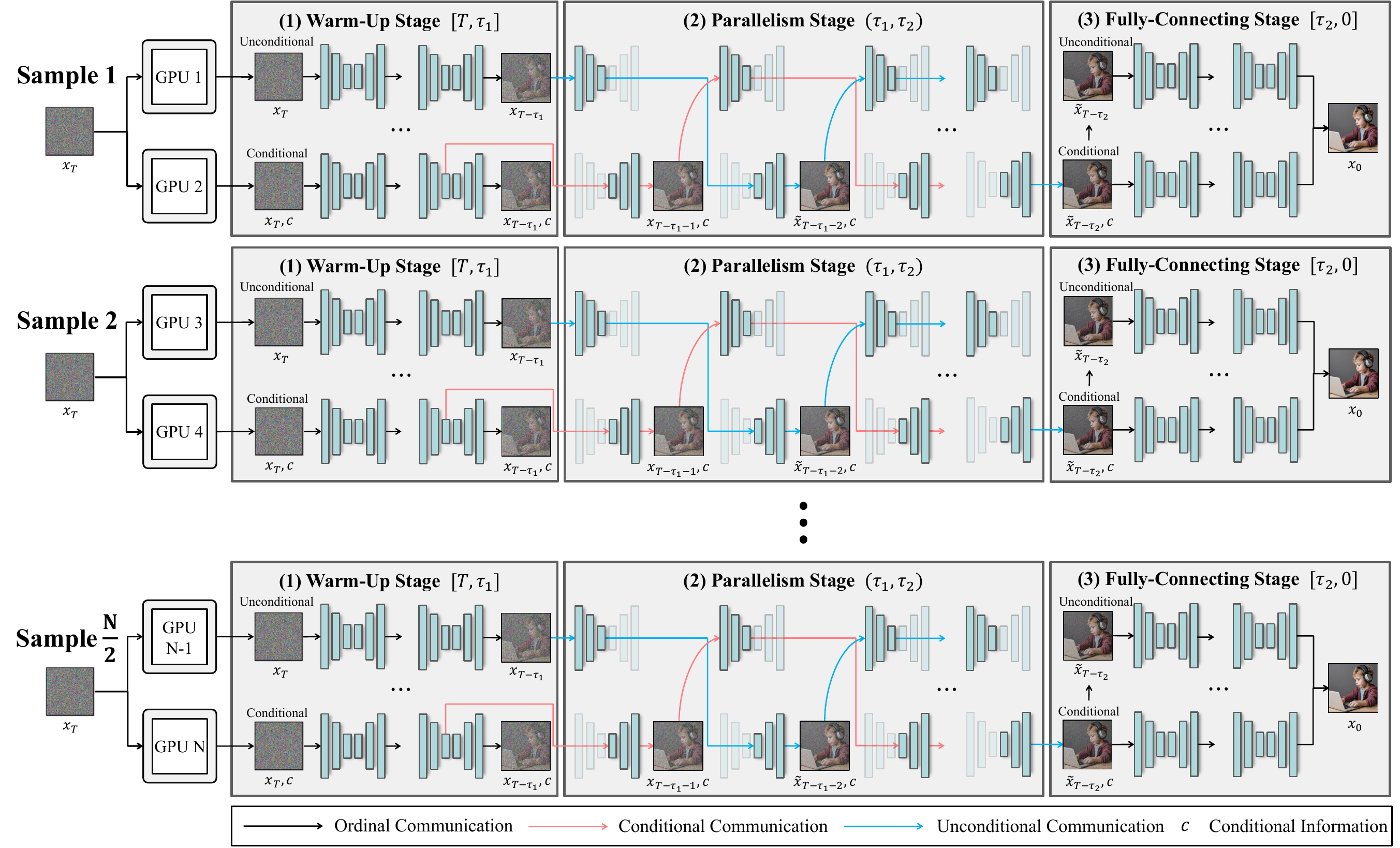}
    \caption{Batch-level extension under $\boldsymbol{N}$\,GPUs configuration.}
    \vspace{5mm}
    \label{fig:app_many_gpu_1}
  \end{subfigure}
  \hfill
  \begin{subfigure}[h]{0.9\linewidth}
    \centering
    \includegraphics[width=\linewidth]{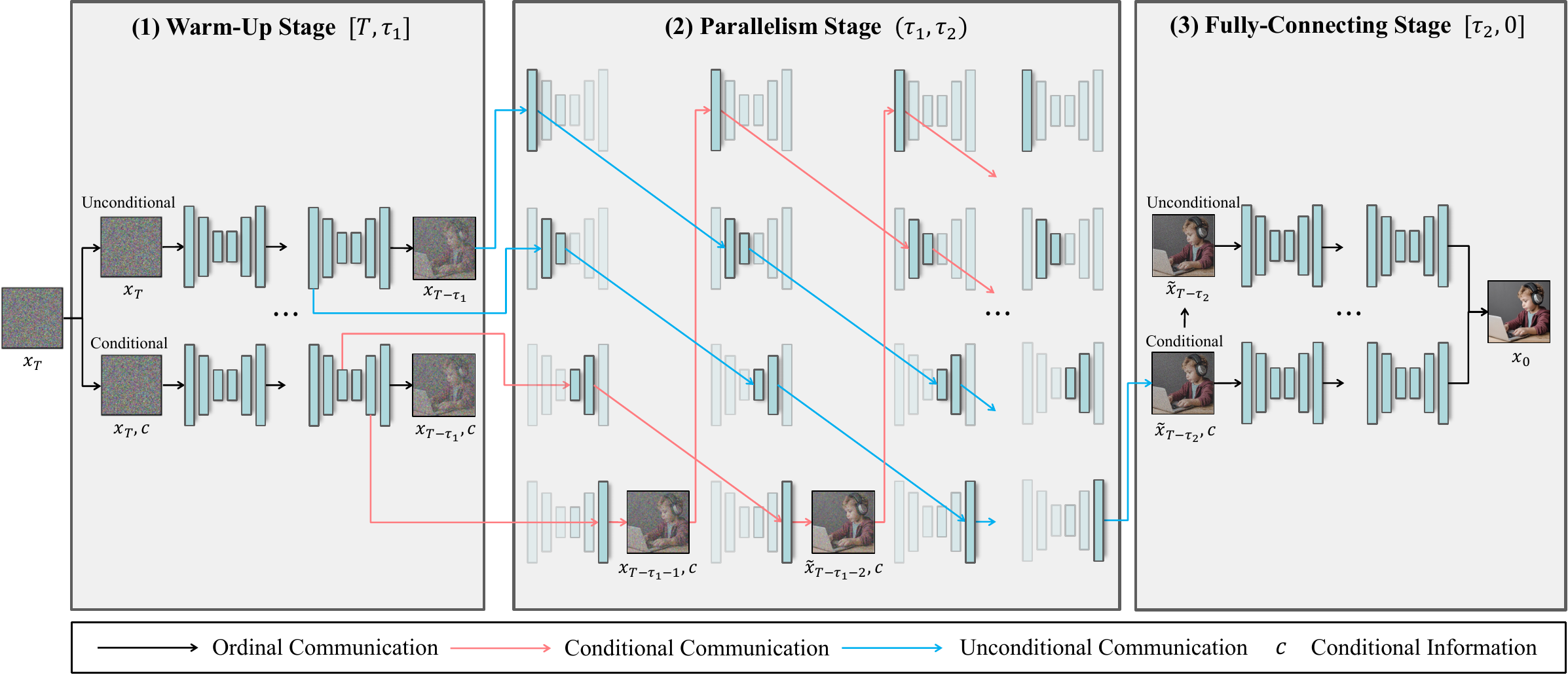}
    \caption{Layer-wise pipeline extension on a 4\,GPUs configuration.}
    \label{fig:app_many_gpu_2}
  \end{subfigure}

    \caption{\textbf{Extensibility to many GPU configurations structures.} This figure illustrates two strategies for scaling the proposed hybrid parallelism framework to larger GPU configurations. These structures demonstrate how the proposed framework naturally generalizes from the 2\,GPUs setting to both batch-level and layer-wise many GPU configurations.}

  \label{fig:app_many_gpu}
\end{figure*}

Figure~\ref{fig:app_many_gpu} presents two extensibility structures that scale the proposed hybrid parallelism framework from the baseline 2\,GPUs setup to many GPU configurations. 

The first structure, shown in Figure~\ref{fig:app_many_gpu_1}, demonstrates the batch-level extension under an $\boldsymbol{N}$\,GPUs configuration. In this scheme, each pair of GPUs collaboratively denoises a single sample while following the three stages hybrid parallelism framework. As a result, the system can generate $N/2$ samples concurrently with $N$\,GPUs, enabling near-linear throughput scaling when multiple samples are produced.

The second structure, shown in Figure~\ref{fig:app_many_gpu_2}, demonstrates the layer-wise pipeline extension on a 4\,GPUs configuration. Here, the denoising network is partitioned into multiple layer-wise segments distributed across devices, allowing the hybrid parallelism strategy to be applied to single-sample generation. While this configuration may exhibit slightly reduced acceleration efficiency and minor quality degradation compared to the batch-level extension, it provides a fine-grained pipeline scheduling mechanism. Importantly, the same structural principles naturally generalize beyond the 4\,GPUs example to arbitrary $N$\,GPUs configurations, demonstrating the flexibility and scalability of the proposed framework.

\onecolumn

\twocolumn 

\section{Implementation Details}
\label{app:implementation_details}
All experiments adopt the DDIM scheduler\,\cite{song2021denoising} with $T=50$ timesteps and generate images at a resolution of $1024\times1024$. Experiments are performed on NVIDIA GeForce 3090 GPUs (24GB each), connected via PCIe Gen3. The adaptive switching parameters are set as follows: for {SDXL}, we use $L=12$, $g_{\text{slope}}=0.4\times10^{-3}$, $k=5$, and $\tau_{\text{cap}}=15$; for {SD3}, we set $L=15$, $g_{\text{slope}}=0.1\times10^{-3}$, $k=5$, and $\tau_{\text{cap}}=40$.

\section{Quantitative Results on the Parallelism \\ Interval $\boldsymbol{k}$}
\label{app:effect_of_k_sec}

\begin{table}[t]
  \centering
  \small
  \setlength{\tabcolsep}{2.2pt}
  \renewcommand{\arraystretch}{1.1}
  \resizebox{\columnwidth}{!}{
  \begin{tabular}{@{}cccc@{}}
    \specialrule{1.0pt}{0pt}{0pt}
    \multirow{2}{*}{\hspace{5pt}\textbf{Parallelism Interval $k$}}\hspace{5pt} &
    \multirow{2}{*}{\textbf{Latency (s)} $\downarrow$} &
    \multirow{2}{*}{\textbf{Speed-Up} $\uparrow$} &
    \textbf{FID} $\downarrow$ \\
    & & & \textbf{(w/ Orig.)} \\
    \midrule
    $k$=5     &  7.12  &  2.31$\times$  &  \textbf{4.100}   \\
    $k$=10    &  6.89  &  2.39$\times$  &  5.942           \\
    $k$=20    &  6.44  &  2.56$\times$  &  7.966            \\
    $k$=30    &  \textbf{5.94}  &  \textbf{2.78$\times$} & 9.191 \\
    \specialrule{1.0pt}{0pt}{0pt}
  \end{tabular}%
  }
  \caption{\textbf{Effect of speed-quality trade-off across different parallelism intervals $\boldsymbol{k}$.} All experiments are conducted on the SDXL model at 1024$\times$1024 resolution with various parallelism intervals.}
  \label{tab:effect_parameter_k}
  \vspace{-3mm}
\end{table}
\noindent Table~\ref{tab:effect_parameter_k} summarizes the numerical values corresponding to the speed--quality trade-off illustrated in Figure~\ref{fig:k_pareto}. As described in the Section~\ref{sec:sensitivity_analysis}, smaller parallelism interval $k$ preserve higher fidelity, whereas larger $k$ values yield powerful acceleration. The table provides concrete measurements that reflect this trade-off, confirming the same trend observed in the pareto frontier visualization.

\newpage

\onecolumn 
\section{Additional Qualitative Results}
\label{app:additional_qualitative}

\begin{figure*}[h!]
  \centering
  \includegraphics[width=0.95\linewidth]{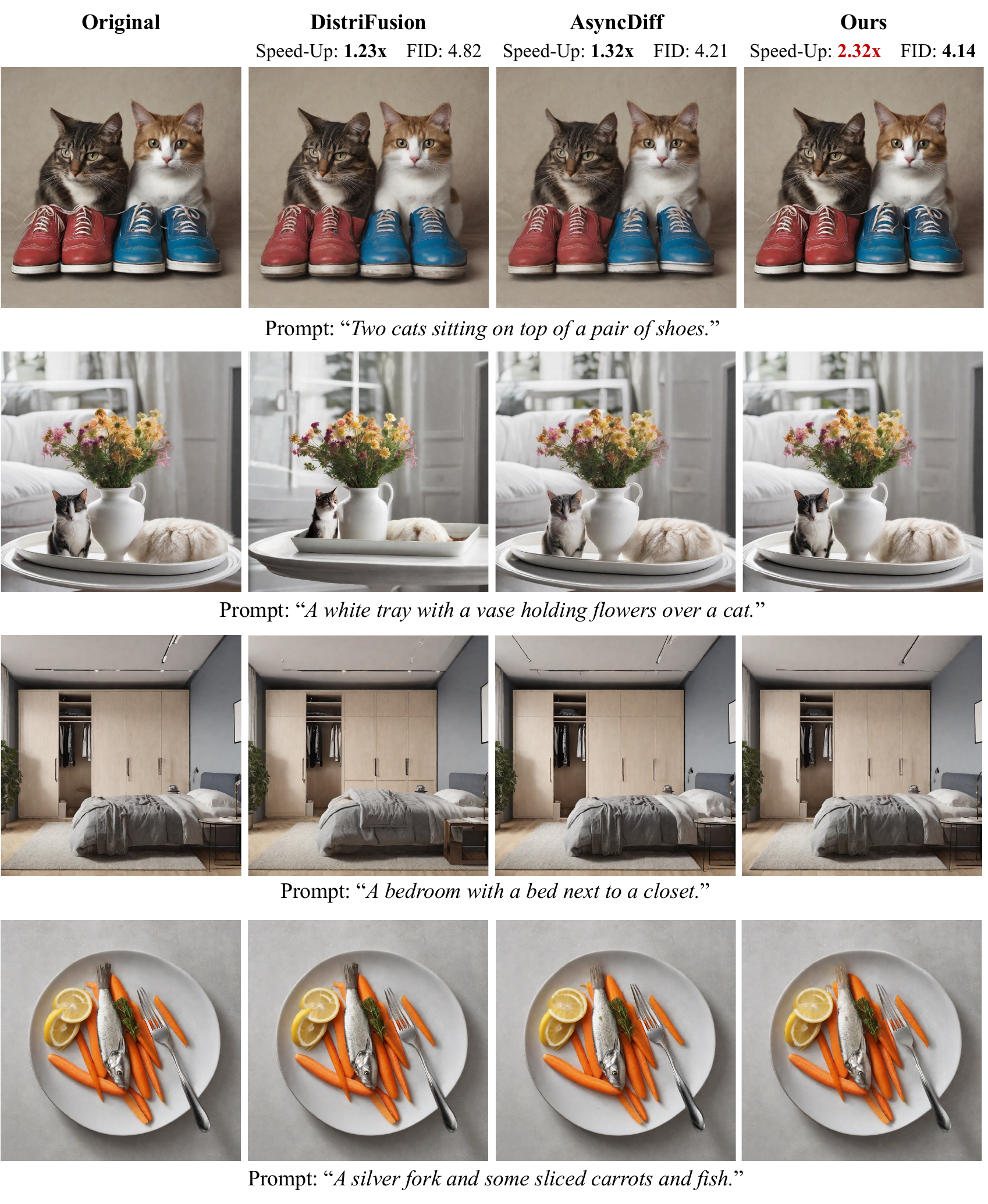}
  \caption{\textbf{Additional qualitative results of the main experiments.} We compare 1024$\times$1024 image generations from the SDXL model. Our method achieves the best acceleration and FID performance, while producing visuals most similar to the original.}
  \label{fig:app_additional_qualitative}
\end{figure*}

\twocolumn 

\newpage

\onecolumn 
\section{Qualitative Comparion Results via Different $\boldsymbol{k}$}
\label{app:qualitative_different_k}

\begin{figure*}[h!]
  \centering
  \includegraphics[width=0.93\linewidth]{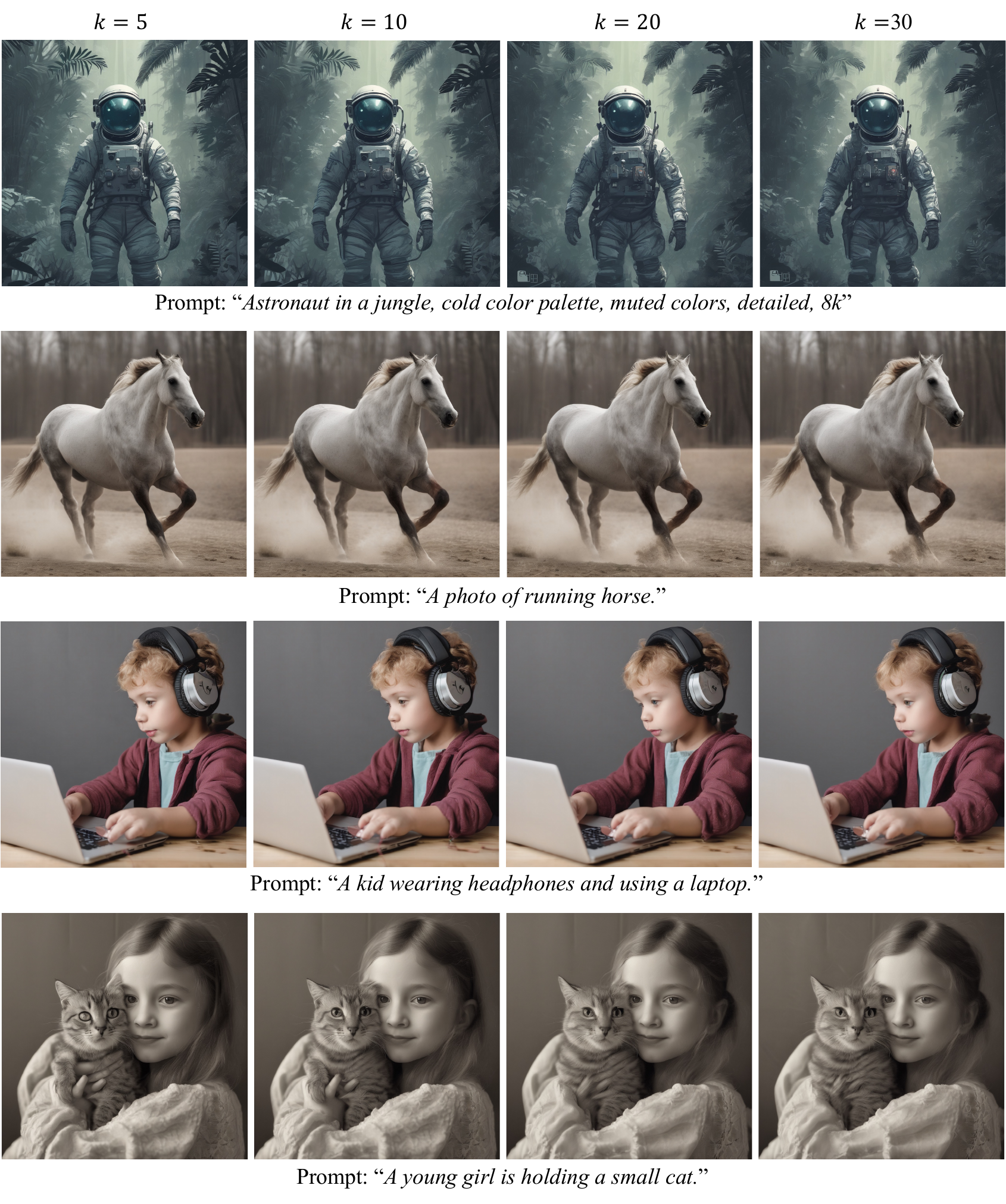}
    \caption{\textbf{Additional qualitative comparisons across different $k$ values.} 
    We compare 1024$\times$1024 image generations from the SDXL model across various parallelism intervals. Smaller $k$ values preserve higher visual fidelity, whereas larger $k$ gradually reduce local detail due to the extended parallelism window. Although the overall appearance remains similar, fine-grained conditional attributes become subtly blurred as $k$ increases.}
  \label{fig:app_additional_qualitative_k}
\end{figure*}

\twocolumn

\end{document}